\documentclass{article}

\usepackage{arxiv}

\usepackage[utf8]{inputenc} 
\usepackage[T1]{fontenc}    
\usepackage[hyphens]{url}            
\usepackage{hyperref}       
\usepackage[hyphenbreaks]{breakurl}
\usepackage{booktabs}       
\usepackage{amsfonts}       
\usepackage{nicefrac}       
\usepackage{microtype}      
\usepackage{lipsum}		
\usepackage{graphicx}
\usepackage[square,numbers]{natbib}
\usepackage{doi}
\usepackage{titlesec}
\titleformat{\subsubsection}[runin]
{\normalfont\normalsize\bfseries\boldmath}{}{0em}{}

\begin{document}
\title{Explanatory Pluralism in Explainable AI}

\author{Yiheng Yao\\Philosophy-Neuroscience-Psychology Program,\\Washington University in St. Louis, St. Louis, MO, USA\\\texttt{yaoyiheng@wustl.edu}}

\date{}

\maketitle              

\begin{abstract}
The increasingly widespread application of AI models motivates increased demand for explanations from a variety of stakeholders. However, this demand is ambiguous because there are many types of `explanation' with different evaluative criteria. In the spirit of pluralism, I chart a taxonomy of types of explanation and the associated XAI methods that can address them. When we look to expose the inner mechanisms of AI models, we develop Diagnostic-explanations. When we seek to render model output understandable, we produce Explication-explanations. When we wish to form stable generalizations of our models, we produce Expectation-explanations. Finally, when we want to justify the usage of a model, we produce Role-explanations that situate models within their social context. The motivation for such a pluralistic view stems from a consideration of causes as manipulable relationships and the different types of explanations as identifying the relevant points in AI systems we can intervene upon to affect our desired changes. This paper reduces the ambiguity in use of the word `explanation' in the field of XAI, allowing practitioners and stakeholders a useful template for avoiding equivocation and evaluating XAI methods and putative explanations.

\end{abstract}
\section{Introduction}

There is no doubt that we should be exacting in our demand on explanations for the outputs, functioning, and employment of AI models, given that they are increasingly implicated in decision making that impact humans with potentially undesirable outcomes \cite{RUDIN}. However, just what we mean by ‘explanations’ in the field of Explainable AI (XAI) is currently unclear \cite{EXPLAININGEX,ViloneLongo}. What is clear is that many different stakeholders have different constraints on the explanations they want from the field \cite{LANGER2021103473}. There is a pressing danger that what explains appropriately and sufficiently is lost in translation from stakeholders to practitioners, and vice versa. In other words, even if the General Data Protection Regulation (GDPR) strongly enforces\footnote{Whether a `right to explain' exists has been debated on the basis of just what explanation is requested by the GDPR \cite{10.1093/idpl/ipx022,GDPR-NOEXPLAIN}. This debate in the literature further highlights the urgency of the present discussion to prevent possible equivocation of the different types of explanation.} that explanations be given when decisions are contested, it is a pyrrhic victory if there are no clear evaluative criteria on the explanations given or worse, that an inappropriate set of evaluative criteria is used to determine which explanation stands as an admissible one.\\

Therefore, when explanations are requested from AI models, and when explanatory demand is placed on the field of Explainable AI, we should first ask some important questions. Why do we ask for ‘explanations’ rather than something else to fulfill the desired goal in posing such a request? What objective or purpose is the explanation supposed to serve in a given context? How should we judge whether a given ‘explanation’ satisfied those objectives or purposes?\\

Without such clarificatory questions, we run the danger of talking past each other in developmental efforts in XAI and in stakeholder’s desiderata for the explanatory products of the field. As astutely noted by Mittelstadt, Russell and Wachter: “many different people …, are all prepared to agree on the importance of explainable AI. However, very few stop to check what they are agreeing to” \cite{EXPLAININGEX}. Langer et.al agree that more clarity is required: “Consistent terminology and conceptual clarity for the desiderata are pivotal and there is a need to explicate the various desiderata more precisely" \cite{LANGER2021103473}. Indeed, going forward, practitioners would benefit from clarity on the requirements for explanations, and stakeholders would benefit from clarity on the limits of explanatory methods produced by the field which would improve their choice of methods to employ.\\

Much recent work in XAI investigates just what are the explanatory demands placed onto XAI by way of analyzing social-psychological constraints \cite{MILLER20191,EXPLAININGEX,PRAGMATICXAI}, how explanations function in the law \cite{DOSHIVELEZ,HACKERTORT}, identifying stakeholders and their desiderata \cite{LANGER2021103473}, and philosophical treatments of explanatory methods \cite{OHARA2020105474,PRAGMATICXAI}. These reviews correctly identify that explanations have a distinct social dimension as a process rather than purely as a product or text \cite{LOMBROZO-EX,MILLER20191,OHARA2020105474}; that explanations should be contrastive, selective, and non-statistical in their content \cite{MILLER20191,EXPLAININGEX}; and that within a social context, explanations of model output do not suffice in isolation \cite{OHARA2020105474}.\\

However, talk of explanations in XAI have remained monolithic. In contrast, the stance I would like to present and defend in this paper is one of Explanatory Pluralism in XAI: the notion that there are many different types of explanation requested from the field for which have different effective treatment by means of methods (what we should produce) and different explanatory powers by range of application contexts (where we can use them). The primary contribution of this paper is a taxonomy, as illustrated in Figure 1, distinguishing the different types of explanation along a Mechanistic-Social axis and a Particular-General axis by identifying the different types of intervention they target. Furthermore, the paper will organize present methods with more specific language introduced using this taxonomy while avoiding the loaded term ‘explanation’.\\
 
\begin{figure}[ht]
\centering
\includegraphics[width=0.8\textwidth]{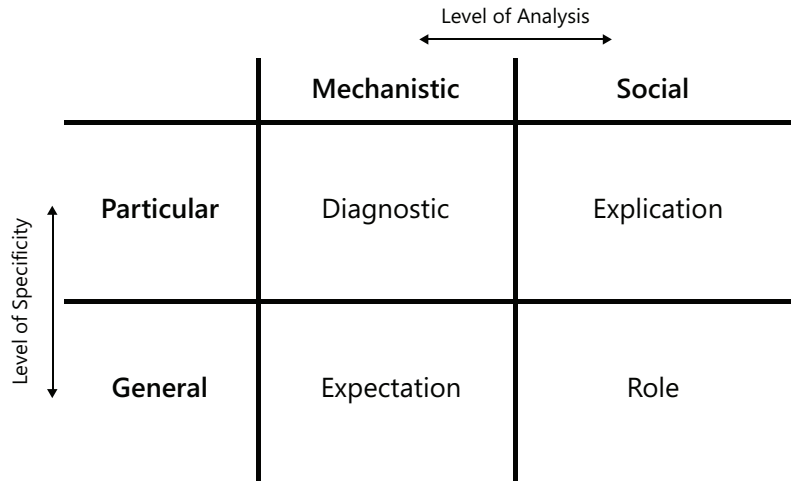}
\caption{Evaluative Taxonomy proposed by this paper to categorize and distinguish the different types of explanation asked for and produced by XAI.} \label{fig1}
\end{figure}

The idea that there are different explanations requested from XAI is not new \cite{OHARA2020105474,ViloneLongo}. However, organization of different explanatory methods have mostly been done in descriptive terms \cite{ViloneLongo}. In this paper, I present a taxonomy based on evaluative terms. XAI methods find membership in the proposed taxonomy by virtue of differences in evaluative criteria rather than differences in descriptive characteristics (when they are assessed -- ex-ante/post-hoc, generality of application -- agnostic/specific, output format, input data, or problem type \cite{ViloneLongo}). By aligning XAI methods with what interventions they target, the success of each method can then be evaluated on the effectiveness of different interventions. The explanatory pluralistic view is also non-reductive, meaning that each category of explanation thus organized do not subsume other categories even though dependency relations may exist between them. I justify my organization of the taxonomy by appeal to recent work on the nature of scientific explanation in the Philosophy of Science, specifically Woodward’s manipulationist account of causation \cite{sep-causation-mani}, Craver’s mechanistic account of scientific explanations \cite{CRAVER-MECHANISM}, and causal relevance \cite{MOREDETAILS}. This normative, philosophically grounded taxonomy serves to specify more clearly what the word `explanation' means in different contexts.\\

I begin by reviewing in Section 2 the diverse explanatory demands for explainability in AI, emphasizing what we are supposed to explain and what we think explanations will help us to achieve. Next in Section 3, I provide relevant contemporary philosophical background drawing from the rich literature in Philosophy of Scientific Explanations to motivate the organization in my proposed taxonomy. In Section 4, I derive and define the Mechanistic-Social, Particular-General taxonomy illustrated in Figure \ref{fig1}. Furthermore, in Section 5, I take a pragmatic interventionist stance and organize present methods in XAI into each of the four categories identified by the proposed taxonomy, showing how methods in each evaluative category fulfill different explanatory demands. Finally, I conclude with two recommendations and highlight that XAI is not merely a way of looking back and within our models but a way of looking forwards and outwards, a perspective ineliminably involved in the development of truly intelligent systems.

\section{Explanatory Demands}

By ‘explanatory demand’ I mean here what is expected of explanations produced by XAI and more broadly, what are the demands placed on the explanatory products of the field (methods which produce explanations). I will anchor my review in the papers by Tim Miller \cite{MILLER20191}, Kieron O’Hara \cite{OHARA2020105474}, and Langer et.al \cite{LANGER2021103473}, organized into three broad areas: social-psychological, social-contextual, and functional (exemplified by stakeholder desiderata). Lastly, I will present what is expected of explanations from the law as examined by Doshi-Velez et.al \cite{DOSHIVELEZ} and highlight some regulatory requirements from the recently proposed Harmonized Rules on AI by the EU \cite{eu-Harmonized}.

\subsubsection{Social-Psychological Demand}
Drawing from Lombrozo’s work on the structure and function of explanations examined as a psychological phenomenon \cite{LOMBROZO-EX}, Miller elucidates some key considerations we expect from explanations when they are given to humans \cite{MILLER20191}: 1) explanations as a social process aim to render something understandable by transferring knowledge between an explainer and explainee; 2) presentation of causes in contrastive terms is preferred; 3) causes cited within an explanation is selective and does not represent the full and complete set of causes; 4) statistical generalizations alone are unsatisfying. In treating explanations as not mere static products but a process that involves social agents, we highlight one important feature of explanations: they elicit understanding (in humans). It is crucial to note that fulfilling this goal sets evaluative criteria that are dependent not upon the content of explanations, whether they do in fact relate to what is explained, but upon how the relevant information is packaged and presented and whether its delivery improves understanding. Put in another way, it is about what makes the light bulb go off in our head, however we reach for the switch\footnote{Craver, 2021, personal communications}. One way of noticing this point is by observing the role of idealized models in science. We do not start teaching with relativity and quantum mechanics but often start by introducing Newtonian physics and constrain our approximate models within some limits such as slow speeds and large sizes. Although such idealized models do not veridically reflect the structure of the world, they lend themselves to better understanding. Of course, we can and should impose the additional constraint that the content of the explanation accurately reflects the underlying causal structure \cite{Craver2014-ONTIC,PINCOCKIDEALMODELS}. However, the important point is that we have both a factivity criterion and an understanding criterion that can be evaluated independently of one another \cite{PRAGMATICXAI}.

\subsubsection{Social-Contextual Demand}
When we employ AI models to aid humans in making decisions or to produce outputs that impact humans, we need to situate the AI as part of a larger social context. O’Hara notes that AI models do not have decision-making power in and of themselves. Administrators can choose to intervene upon systems, and how the output is acted upon is distinct from the mechanisms of the AI model that generated it \cite{OHARA2020105474}. As such, when we ask for explanations regarding decisions ‘made’ by AI models, we ought to include relevant details of where such a model is situated in the broader social context surrounding its usage. 

\subsubsection{Functional Demand}
Langer et.al compiles a comprehensive assessment of the different stakeholders who are interested in seeking explanations from AI models \cite{LANGER2021103473}: 1) Users seeking \textit{usability} and \textit{trust}; 2) Developers seeking \textit{verification} and \textit{performance}; 3) Affected parties seeking \textit{fairness} and \textit{morality/ethics}; 4) Deployers seeking \textit{acceptance} and \textit{legal compliance}; and 5) Regulators seeking \textit{trustworthiness} and \textit{accountability}. The type of explanation that prove useful to developers of AI models for the purpose of debugging or improving model accuracy would look very different than that which a non-expert user may request for understanding how their personal data is used, precisely because they serve such different purposes. Therefore, it would be insufficient to simply claim that `explanations' help in all these diverse cases, we need to further specify what type of explanation would help by clarifying the explanandum (what is to be explained).

\subsubsection{Legal Demand}
Explanations are of value in legal settings for holding AI systems accountable by ``exposing the logic behind a decision" and to ``ascertain whether certain criteria were used appropriately or inappropriately in case of a dispute" \cite{DOSHIVELEZ}. In their review, Doshi-Velez et.al also note that explanations will be requested only when they ``can be acted on in some way" \cite{DOSHIVELEZ}, highlighting the cost-benefit trade-off in generating explanations. In addition, the authors note that further explanations may be demanded ``even if the inputs and outputs appear proper because of the context in which the decision is made" \cite{DOSHIVELEZ}. Here, three demands on explanations are emphasized. They must: 1) identify contributing factors to the output; 2) identify actionable factors specifically; and 3) attend to the context in which the AI system is deployed to make decisions and take actions. In addition, the proposal for Harmonized Rules on AI in the EU sets additional requirements on employing ``high-risk'' AI systems intended to be used as ``a safety component'' \cite{eu-Harmonized}. The intent of the proposal echoes that of Article 22 in the GDPR that placed restrictions on automated decisions ``which produces legal effects ... or similarly significant affects'' on humans subjected to such decisions \cite{eu-GDPR}. In both cases, regulators are interested in identifying AI systems that play a significant role in impacting humans and place additional restrictions on their usage. Furthermore, the newly proposed Harmonized Rules on AI additionally introduce a “Technical Documentation” requirement in Article 11(1) for fielding such “high-risk” AI systems \cite{eu-Harmonized}. This document as described in Annex IV includes a comprehensive list of information such as “how the AI system interacts or can be used to interact with hardware or software that is not part of the AI system itself, where applicable” (Annex IV 1(b)), “what the system is designed to optimize for and the relevance of the different parameters” (Annex IV 2(b)), and “metrics used to measure accuracy, robustness, cybersecurity” (Annex IV 2(g)) \cite{eu-Harmonized}.

\subsection{Fulfilling Disparate Explanatory Demands}

Explanations are sought for in a multitude of situations, with a diverse set of goals and expectations as reviewed in this section. Considering the importance of explanations in ensuring the responsible usage of AI systems, there is a pressing need to evaluate the quality of explanations given. However, what constitutes as a meaningful explanation differs to the different stakeholders involved. Therefore, we should first acknowledge the plurality of explanations and distinguish between the different types of explanation so we can develop the appropriate evaluative criteria and methods to address the different requests for meaningful explanations. In the next section, I will appeal to recent work in the Philosophy of Science on the nature of scientific explanations to show how we can differentiate between requests for explanations by identifying the relevant level of change in the AI model we wish to affect using the notion of causal relevance.

\section{Scientific Explanations}

Much has already been said on the nature of explanations, especially what are good explanations in the sciences \cite{sep-scientific-explanation}. One point of agreement between scientific explanations and past work on the nature of explanations in XAI is that explanations should unveil causes \cite{Craver2014-ONTIC,MILLER20191}. However, evaluating the quality of explanations based on the amount of causes they identify or how many why-questions they can answer is insufficient \cite{MOREDETAILS}. As previously acknowledged, explanations should further be selective \cite{MILLER20191,EXPLAININGEX}. I appeal to recent developments in the Philosophy of Science to state more clearly how we should be selective with our explanations.


\subsection{Manipulationist Account of Causation}

Firstly, just what is this notion of a `cause'? The manipulationist account of causation put roughly is that: X causes Y if manipulating X changes the value of Y or its probability distribution. Put in another way, ``causal relationships are relationships that are potentially exploitable for purposes of manipulation and control" \cite{sep-causation-mani}. Furthermore, Woodward introduces a stability constraint in evaluating which cause is more suitable given some effect Y \cite{WOODWARDSTABILITY}. Under the stability constraint, causal relationships which ``continue to hold under a `large' range of changes in background circumstances" \cite{WOODWARDSTABILITY} should be preferable. This may be a driver for the social-psychological demand for explanations presented in contrastive terms. The larger the range of counterfactuals identified under which the causal relationship holds, the more inclined we may be in accepting the identified cause.

\subsection{Mechanistic Account of Scientific Explanations}

Craver builds upon this notion of causes as manipulable relationships, or points of intervention\footnote{In this paper, I sometimes use the term interventions in place of manipulable relationships. The difference between a manipulable relationship and an intervention \cite{sep-causation-mani} is a subtle one that does not affect my arguments.}, to develop a mechanistic account of explanations for cognitive neuroscience. In this account, explanations describe mechanisms which are ``entities and activities organized such that they exhibit the \textit{explanandum} phenomenon'' \cite{CRAVER-MECHANISM}, where entities are the components or parts in a mechanism and activities are causes in the manipulationist sense. Three elements of the mechanistic account will be helpful for explicating different types of explanations in XAI: 1) explanations reveal the relevant causal organization of the explanandum at multiple levels; 2) different explanations given at different levels of realization are non-reductive; and 3) relevant causes are those which make a difference to the effect contrast asked for. In summary, the causal organization revealed by different explanations identify different relevant relationships which can be exploited for purposes of manipulation and control.

\subsubsection{Levels of Explanation}
Within a mechanism, activities and components in a lower level are organized to realize higher level activities or components \cite{CRAVER-MECHANISM}. Furthermore, such levels are ``loci of stable generalizations'' \cite{CRAVER-MECHANISM} in the sense that the behavior of components within each level are regular and predictable \cite{CRAVER-MECHANISM}. When we ask for explanations of a mechanism, we can attend to different levels to identify different stable generalizations we are interested in. For example, when we examine an AI model, we may be interested in the behavior of a range of components located at different levels of realization such as the training hyperparameters, model architecture, and optimization function.

\subsubsection{Non-Reductive}
Since there are stable generalizations of mechanisms that are not true of the arrangement of components that realize them \cite{CRAVER-MECHANISM}, there are different causally relevant sets of components at different levels of realization. Explanations of general AI model behavior such as identifying what rules they follow in processing patterns of input features need not necessarily be better substituted with explanations of particular AI model processes that led to an output. The latter may add further details to the former but without situating such details within a higher level, it would be difficult to ascertain similar generalizations of model behavior. An analogy is that to explain the functioning of a program, we need not reduce our explanations to the movement of electrons in the CPU although such movement does realize the program under question at a lower level.

\subsubsection{Causal Relevance}
The notion of causal relevance stems from the non-reductive nature of levels of explanation as considered above. Causes which are relevant to an explanation should identify ``the `differences that make a difference.'" \cite{MOREDETAILS} When we seek explanations, inherent within our request is some class of effect contrast we are attending to. For example, when asking why an AI model classifies images in some way, we may attend to the particular relevance of some subset of features versus others as our contrast class or the distribution of labels over one dataset versus another. To effectively address the request for explanations, we should provide causes relevant to bringing about changes in the requested contrast class. The two ways we answer the question why an image is classified the way it is identify different points of intervention at different levels, so as to change the model behavior in different ways. By attending to particular feature relevance, we target changes in the model's output for a range of similar inputs. By attending to label distribution, we target changes in the model's classification behavior when given different datasets. It is therefore crucial to clarify what is the desired effect contrast so we can provide an appropriate explanation. The notion of an explanation revealing relevant causes at the appropriate level affords us a way to demarcate different types of explanation by identifying different levels of realization, different effect contrasts, and different points in AI systems where we can intervene.

\section{Pluralistic Taxonomy}
With the need to identify the desired effect contrast at different levels of realization as discussed in Section 3.2, I derive my proposed pluralistic taxonomy by augmenting David Marr’s famous Three-Levels of Analysis widely applied in cognitive psychology and originally tailored for the biological visual system \cite{MARR-LEVELS}. Furthermore, drawing inspiration from the taxonomy of Scientific Explanations introduced by Hempel that distinguishes between Particular Facts or General Regularities and Universal Laws or Statistical Laws \cite{HEMPELASPECTS}, I arrive at a taxonomy similarly based on a Specific-General axis that additionally considers the augmented levels of analysis along a Mechanistic-Social axis.

\subsection{Three Levels of Analysis (Plus One)}

\begin{figure}[ht]
\centering
\includegraphics[width=\textwidth]{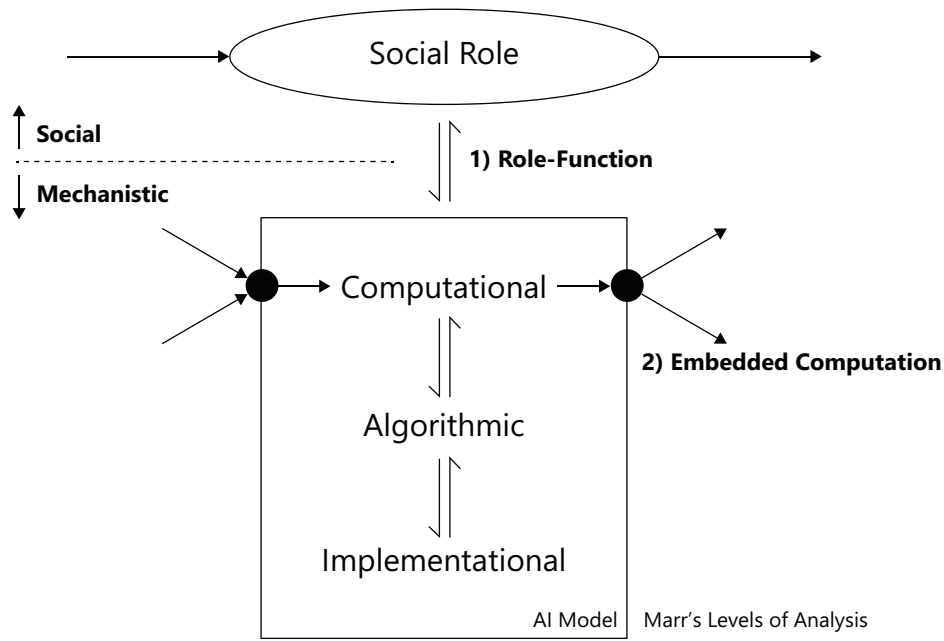}
\caption{David Marr's Three Levels of Analysis for information processing systems adapted for AI model employment within social contexts.} \label{fig2}
\end{figure}

Neuroscientist David Marr introduced three levels of analysis to aid with understanding information processing systems \cite{MARR-LEVELS}: the Computational level (the goal or problem solved); the Algorithmic level (processes and mechanisms used to solve said problem); and the Implementational level (physical substrate used to realize such mechanisms). Mapped onto AI terminology: 1) at the Computational level we can describe our models based on what task it attempts to perform (image classification, text-based summary generation, function minimization, etc.); 2) at the Algorithmic level we can describe what architecture is employed to solve this task (LSTM, RNN, GMM, etc.); and 3) at the Implementational level we can specify what are the hyperparameters that instantiate this particular model and the hardware we use to run it (TPU hours used, Bayesian Optimization value/acquisition functions used, etc.).\\

However, limiting analysis to the aforementioned three levels of analysis would be insufficient as XAI is specifically interested in types of computation and AI models that are used in some way that influences human decision making, or its outputs impact humans in some other way. For example, we are not typically interested in an isolated NPC (non-playable character) AI within some computer game which may similarly be decomposed into these levels of description and analysis. As such, highlighted in Figure \ref{fig2}, we need to acknowledge that: 1) the AI models employed realize some Social Role, and 2) the AI model is embedded in additional computation surrounding its usage. Rarely do we have an AI model for which the input is statically specified, and its output directly used \cite{OHARA2020105474}.\\

By Social Role, I mean to draw attention to the set of societal expectations surrounding decisions made in the context of application \cite{SOCIAL-ROLE}. One may question the authority, ethics, and suitability of the AI model’s (or the system's in general) continued employment in such a position that impacts humans or human decision-making. It is one question to ask whether an AI model is functioning as designed and an entirely separate question to ask whether the AI model thus designed could satisfactorily play the role we cast it in. The latter requires that we look outwards to position the AI model within its broader social context and identify whether it satisfies what is expected of such roles they may come to occupy. Granted, part of the difficulty here is that social expectations are typically not explicit\footnote{Interestingly, there has been research to determine the social norms surrounding trolley-like decision problems in the context of an imminent car crash \cite{MIT-TROLLEY}.} \cite{DOSHIVELEZ} and the systems we have for establishing suitable membership in social roles are tailored for human agents\footnote{Non-human animals are not recognized as legal persons and cannot stand in courts \cite{NONHUMAN-COURT}. Can an AI system stand in court as a defendant?}.\\

The addition of a Social level to Marr’s three levels of analysis emphasizes the point that AI models do not operate in isolation, at least not the ones interesting to XAI. No matter how brightly we illuminate the mechanistic details within the AI model, no matter how transparent our algorithms are \cite{ALGO-TRANSPARENCY}, we are missing a big chunk of the picture if we restrict discussion to analysis of only the Computational, Algorithmic, and Implementational levels.

\subsection{Mechanistic-Social, Particular-General Taxonomy}

In addition to the Mechanistic-Social levels of analysis distinction\footnote{That is not to deny that there may be social mechanisms.}, we may also ask for explanations at different levels of specificity much like how Hempel distinguished between explaining particular facts from general regularities \cite{HEMPELASPECTS}. Here, we distinguish between asking questions pertaining to why a particular output was produced, and what types of output tend to be generated. We are also distinguishing between whether a particular social agent can understand outputs or explanations generated by XAI methods, and whether the usage of the AI model under question fits within the broader social context of application.\\

To be precise in our usage of language and avoid the ambiguous and loaded term ‘explanation’, each category in this taxonomy introduces a distinct term to disambiguate discourse. When we talk about explanations that identify mechanisms within an AI model contributing to particular outputs, we request for and produce Diagnostic-explanations on the matter (Mechanistic/Particular). When we wish to discern the general regularities of an AI model, we request for Expectation-explanations (Mechanistic/General). When we talk about explanations given to humans, we are requesting for Explication-explanations (Social/Particular). Finally, when we ask for justifications of model usage and seek guidance on regulations and policy, we request for Role-explanations which position an AI model within its context (Social/General).\\

The advantage of introducing this distinct terminology is two-fold. Firstly, we can keep separate questions which require different XAI methods to address appropriately and develop evaluative metrics and methods within each category independently. Secondly, we can now talk clearly about the relationship between each of these types of explanation and explanatory methods produced by XAI. Furthermore, adopting the view of explanatory pluralism means that we do not place primacy on any one type of explanation but acknowledge that there many types, each suiting a different context or need. For example, it is not the case that a Role-explanation should always be given, as it would do little to determine whether a particular AI model is actually functioning the way it was designed to. An analogy here is that it is insufficient to ascertain that the person who gave the (incorrect) prescription was a doctor. Rather, we still need to ascertain particular facts of the matter such as whether the doctor made errors in judgment, or employed incorrect diagnostic tools, or whether such tools failed to function correctly which factored into the decision to prescribe the wrong medication. However, it is the case that if we were asking whether it was acceptable that this particular person gave someone else a prescription, we determine whether the person under question is a trained doctor or pharmacologist.\\

\begin{figure}[ht]
\centering
\includegraphics[width=0.8\textwidth]{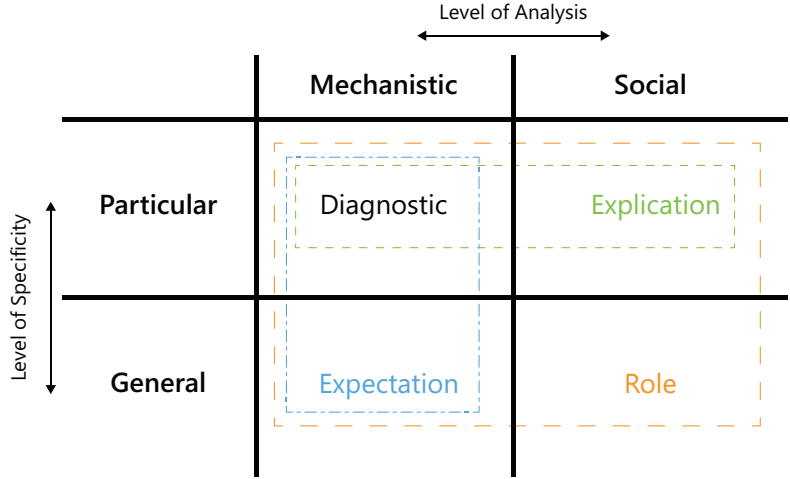}
\caption{Highlighting the dependency relations between the different evaluative categories.} \label{fig3}
\end{figure}

This prescription analogy hints at the dependency relations, as highlighted in Figure \ref{fig3}, between different types of explanation in the proposed taxonomy. When we give explanations that are explicable to human receivers fulfilling the set of social-psychological constraints, we also need to ensure that what we explicate match the mechanisms that produced the object of explication. In other words, as noted by Rudin, there is a worry that explanations produced may not match what the model computes \cite{RUDIN}. Therefore, it is important to establish that whatever explanations that are explicable in terms of being contrastive, selective, and non-statistical (criteria noted in Section 2) be nonetheless grounded with suitable and accurate Diagnostic-explanations unveiling the relevant mechanisms in the model under scrutiny. Similarly, even if we were to use an Interpretable model with provable bounds which we can generate Expectation-explanations for, we still need to make use of diagnostic methods to verify that the model is functioning correctly. Moreover, simply putting a model for which we have certain bounded expectations on the table does not make its output immediately explicable, although having prior expectations might mean that model outputs lend themselves to easier explication. Expectation-explanations provided still need to fulfill a set of explicability criteria to be understandable to the target audience. Finally, to determine whether a model fits social expectations for the role that it occupies in its social context, we may require that it both be understandable to humans interacting with it and that we can draw generalizations around its function. But an explanation that is both explicable and based on an Interpretable model architecture may, however, still fail to identify and position the model within its social context and thus, fail to be a suitable Role-explanation.

\section{Pragmatic Interventionist Stance}

We can now position XAI methods within this pluralistic taxonomy by identifying the knobs and levers that we should manipulate to affect our desired effects (fulfillment of desiderata). In other words, the different categories differ in where we intervene upon our system to exact the desired changes. Taken together with the notion of causal relevance, the pragmatic interventionist stance, that explanations help us uncover relevant causes which identify manipulable relationships, affords us a unified way of categorizing XAI approaches.

\subsection{Organizing Present XAI Methods}

\begin{figure}[ht]
\centering
\includegraphics[width=0.8\textwidth]{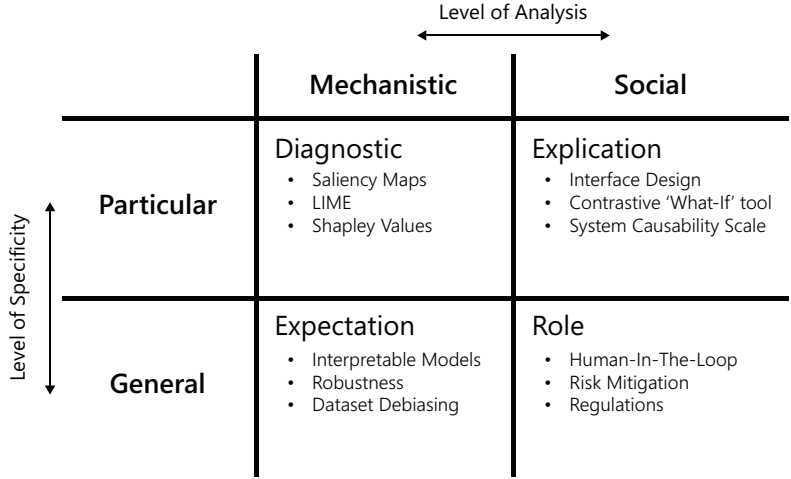}
\caption{Selected examples organized under each category in the proposed taxonomy.} \label{fig4}
\end{figure}

For a more comprehensive review of the methods in XAI, I refer the reader to \cite{ViloneLongo}. In this section, I have chosen some representative examples to illustrate the application of my proposed taxonomy in Figure \ref{fig4}. Within the category of Diagnostic-explanation are Saliency Maps \cite{SALIENCY-RAP}, LIME \cite{LIME}, and Shapley Values \cite{GOOGLE-XAI} which identify particular input features important to affecting the output of models. The Explication-explanation category focuses on techniques to render explanations or model output understandable to humans interacting with the AI model. Such methods may include refining AI model interfaces with Human-Computer-Interaction (HCI) research \cite{HCI-example}, Google’s AI Explanations “What-If” tool \cite{GOOGLE-XAI-WHATIF} to present feature relevance in contrastive terms, or by using the System Causability Scale \cite{causability} to measure the extent to which explanations generated were understandable. The Expectation-explanation category includes methods that focus on identifying and building regularities into models \cite{GUARANTEE-UNCERTAINTY,GUARANTEE-SAFE}, ensuring robustness against adversarial attacks \cite{ROBUSTNEESS-GP}, and avoiding a pattern of output that potentially biases towards inappropriate features \cite{DEBIASING}. Interpretable models by virtue of their architectural attributes allow us to form certain expectations. For instance, the Neural Additive Model architecture uses a linear combination of neural networks to compute classification \cite{NAM}. We can expect that a linear combination will combine each input feature in some weighted additive manner rather than have potentially unexpected interactions between features in high dimensions as deep neural networks typically do. Role-explanation emphasize the social context and embedded nature of AI models. By explicitly including humans in the process of decision-making and training, the consideration of Human-In-The-Loop is three-fold: 1) humans may be required to review AI model decisions to comply with regulatory constraints; 2) humans can augment AI models with expertise and skills that AI models currently do not possess \cite{medical-ai}; and 3) by including humans within each stage of the AI model, we can better ensure that AI objectives are aligned with human values since such systems open themselves up to more flexible alignment with human preferences \cite{HCI-DESIGN}. Furthermore, risk mitigation protocols, as required for ``high-risk'' AI systems under the proposal for Harmonized Rules on AI \cite{eu-Harmonized}, may identify ways to recover control when the AI system steps outside the boundaries of the role it plays, thereby increasing our trust in the AI system to perform within suitable roles. Finally, the growing body of regulations can help us to clarify what are the roles we envision AI systems can act beneficially within and their associated expectations.


\subsection{Descriptive vs Evaluative Taxonomy}
XAI methods can be categorized as illustrated in this proposed taxonomy by how we should evaluate them based on the sorts of intervention they identify instead of descriptive characteristics. Since causes identified by explanations should be relevant to the effect contrast we wish to affect \cite{CRAVER-MECHANISM,MOREDETAILS,WOODWARDSTABILITY}, we should ask for XAI methods from the appropriate category of interventions. If we wish to examine changes in the model output, we should intervene at the level of a particular trained model asking for Diagnostic-explanations. If we wish to determine the broad guarantees of a model, then we should intervene at the level of the model architecture and ask for Expectation-explanations. If what we ultimately wish for is human understandability, then we should intervene upon the causes that bring about increased understandability, such as the social-psychological considerations outlined in Section 2, and ask for Explication-explanations. Finally, if we wish to better fit the usage of our AI model within its social role, perhaps what we should intervene upon is not the model architecture nor how explicable outputs are, but to involve human controllers and specify their operating procedures or develop regulatory mechanisms surrounding usage of such models and ask for Role-explanations.\\


Therefore, in addition to disambiguating discourse on different XAI methods, the proposed taxonomy also allows stakeholders to identify a match between their desiderata and the methods that should be employed. For example, if we want to “restore accountability by making errors and causes for unfavorable outcomes detectable and attributable to the involved parties” \cite{LANGER2021103473}, what we are looking for will be Diagnostic-explanations that identify particular mechanisms in the model contributing to errors as well as Role-explanations that identify the context within which the model was situated. If we wish for users to “calibrate their trust in artificial systems” \cite{LANGER2021103473}, then we request for Explication-explanations to render model output understandable and Expectation-explanations to identify robustness guarantees.

\section{Conclusion}

In conclusion, this paper presents an evaluative taxonomy that categorizes XAI methods based on the levels of intervention available and acknowledges the plurality of explanations produced. Furthermore, distinct terminology is introduced for each category to disambiguate the types of explanation we mean: Diagnostic, Expectation, Explication, and Role-explanation. This taxonomy is neither complete nor the only such way we can organize different types of explanation. Rather, this paper makes the point that it is useful for us to differentiate between types of explanation and we should do so on the basis of evaluative criteria rather than descriptive criteria. Additionally, future work is encouraged to develop metrics for evaluating XAI methods in each category. In particular, contributions from the social sciences will be crucial in identifying just what we should look for in Explication-explanations and Role-explanations. Nevertheless, we can now answer some of the clarificatory questions posed in the introduction. Why do we ask for `explanations'? Because they allow us to identify relevant points of intervention for the desired effect. Furthermore, with the more specific language introduced, we can better distinguish between the evaluative conditions we wish to impose upon explanations requested. This allows stakeholders to more clearly present their objective, purpose and context under which explanations are sought from XAI. I will end with two recommendations for XAI, reemphasizing the point that rather than looking back upon and within our present models, methods developed in XAI can look forward as a way of advancing the field of AI and should look outwards to situate models within their social context.

\subsection{Limit of Diagnostics}
The first pressing recommendation is to use the more specific term `AI Model Diagnostics' when we talk about explanatory methods that illuminate mechanisms within AI models. It would be prudent to treat present results from the field of XAI that are mere diagnostic tools as such explicitly to avoid confusion and granting such tools too much authority.\\

This difference between Diagnostics and full `explanations' can be illustrated with an analogy to Air Crash Investigations. In the unfortunate case of airplane accidents, the recovery of the plane's Flight Recorder (also known as a black box) is but the first step in forming an investigative report into the accident. The data recorded by flight recorders contain a slice of the plane's flight history, preserving the state of the plane moments before the accident. From this data, investigators may be able to hypothesize what caused the accident by identifying anomalous parameters recorded by the black box. However, in many cases, the causes for airplane accidents do not lie entirely within the plane's state prior to the accident. Rather, the plane exists within the larger context of the flight industry which contains its pilots, maintenance crews, and regulations regarding flight paths and operating procedures. In addition, a key aspect of the final investigative report is to not only identify causes for the accident but recommendations for preventing future accidents from happening \cite{AAIB}. Furthermore, this investigative report also serves to assuage the public of the flight industry's reliability as well as address bereaved families' concerns. In this way, explanations for airplane accidents do not merely contain the causal aspect (which may already exceed the bounds of a plane's black box) but a social aspect of fulfilling the responsibility the flight industry has to its customers.\\

In a similar fashion, our investigations into AI models must not stop at uncovering what's within the black box (AI models), but look beyond and place the model within its social context. But to do so, we do indeed still require transparency into the inner workings of our AI models in order to render their behavior expectable and explicable. Therefore, a more holistic approach to constructing XAI explanatory products may be necessary by incorporating methods from multiple categories within the proposed taxonomy.

\subsection{Ratiocinative AI}
The second long-term recommendation is to identify an additional direction XAI can take that is somewhat distinct from the any of the categories defined in the proposed taxonomy: bake into AI models an awareness of its internal processes. In Rosenberg's critical take on connectionism, \textit{Connectionism and Cognition}, he argues that the ``mere exercise of a discrimination capacity, however complex, is not yet an example of cognition" \cite{ROSENBERG} and identifies connectionist networks (neural networks) as only capable of mere discrimination by following certain rules. In addition to being rule-conforming (rational), Rosenberg argues in his paper that for truly cognizant systems, they should be rule-aware (ratiocinative) as well. Just so, I believe that for us to eventually develop tools that enable fruitful dialogue between humans and our AI models, we would come to imbue our AI models with an awareness of its internal processes.\\

What this awareness should be and how it can be implemented is currently unclear. XAI can contribute by not only identifying points of intervention, but eventually allowing us to reflexively surface these interventions back to the AI to develop truly intelligent, ratiocinative systems. Furthermore, present research in Reinforcement Learning agents also paints a promising path for us to achieve this goal. Building on top of its previous successes, DeepMind's recent MuZero agent is able to learn both the rules surrounding permissible actions within its environment as well as an optimal policy to act within this environment \cite{MUZERO}. MuZero's awareness of internal rules and policies learned through interacting with the environment is built upon similarly opaque deep neural networks. However, the levels at which we can direct it questions and extract explanations appear to be broader than most other current AI models. 

\subsection{On Firmer Grounds}

In closing, the categorization of many present XAI methods as `Diagnostics' and admitting a plurality of explanations, thus noting the insufficiency of any single type of explanation, may be viewed as taking a step back. However, by taking this step back to reign in and clarify some of the expectations we have for present XAI methods, we stand on firmer grounds to take the next leap forward in XAI to produce holistic explanations and ensure the responsible usage of AI in society.

\subsubsection{Acknowledgments} I thank Carl Craver for many fruitful discussions and helpful comments on multiple drafts of this paper, Jin Huey Lee for feedback on an earlier draft, and my anonymous reviewers for their many insightful comments.

\bibliographystyle{splncs04}
\bibliography{refs}

\end{document}